\documentclass{article}
\usepackage{authblk}
\usepackage{graphicx}
\usepackage{tabularx} % extra features for tabular environment
\usepackage{booktabs} % For formal tables

\usepackage{multirow}
\usepackage{adjustbox}
\usepackage{algorithmic}
\usepackage{xurl}
\usepackage{amsmath,amssymb,amsfonts}
\usepackage[ruled, noend]{algorithm2e}
\SetKwComment{Comment}{$\triangleright$\ }{}
\mathchardef\mhyphen="2D % Define a "math hyphen"
\usepackage{xcolor}
\newcommand{\ignore}[1]{}
\begin{document}
\title{Novelty-Driven Binary Particle Swarm Optimisation for Truss Optimisation Problems}
\author[1]{Hirad Assimi}
\author[1]{Frank Neumann} 
\author[1]{Markus Wagner} 
\author[2]{Xiaodong Li} 
\affil[1]{School of Computer Science, The University of Adelaide, Adelaide, 5000, South Australia, Australia}
\affil[2]{School of Computing Technologies, RMIT University, Melbourne, 3001, Victoria, Australia}

\makeatletter
\renewcommand\AB@affilsepx{, \protect\Affilfont}
\makeatother

\affil[1]{firstname.lastname@adelaide.edu.au}
\affil[2]{xiaodong.li@rmit.edu.au}
% \affil[ ]{\textit {\href{mailto:benchmarkingbestpractice@gmail.com}{benchmarkingbestpractice@gmail.com}}}
\setcounter{Maxaffil}{0}
\renewcommand\Affilfont{\itshape\small}

\date{}
\maketitle
\begin{abstract}
Topology optimisation of trusses can be formulated as a combinatorial and multi-modal problem in which locating distinct optimal designs allows practitioners to choose the best design based on their preferences. Bilevel optimisation has been successfully applied to truss optimisation to consider topology and sizing in upper and lower levels, respectively. We introduce exact enumeration to rigorously analyse the topology search space and remove randomness for small problems. We also propose novelty-driven binary particle swarm optimisation for bigger problems to discover new designs at the upper level by maximising novelty. For the lower level, we employ a reliable evolutionary optimiser to tackle the layout configuration aspect of the problem. We consider truss optimisation problem instances where designers need to select the size of bars from a discrete set with respect to practice code constraints. Our experimental investigations show that our approach outperforms the current state-of-the-art methods and it obtains multiple high-quality solutions.
\end{abstract}

\section{Introduction}
Trusses are used in the construction of bridges, towers ~\cite{hasancebi2007optimization,rao1995optimum}, aerospace structures~\cite{seber2011multidisciplinary} or in robots~\cite{finotto2013hybrid} and they carry applied external forces on nodes to support structures. Truss topology optimisation can be expressed as a combinatorial optimisation problem. The goal of topology optimisation is to decide whether to include or exclude necessary bars so that the structure's weight is as light as possible while adhering to structural constraints. 

The ground structure method~\cite{topping1983shape} gives the excessive potential connections between nodes and the preliminary nodal positions in the design space, representing an upper bound in the combinatorial search space of the topology. Therefore, the optimal design is found as a subset of ground structure. Truss optimisation also includes size and shape optimisation, where size optimisation determines the optimal sizing of active bars in the design and shape optimisation finds the optimal nodal positions to determine the truss layout. 

Achieving an overall minimum weight of a truss is the most common objective to save costs and improve efficiency \cite{brooks2017undeflected} in truss optimisation problems based on ground structure. Truss optimisation is challenging due to its nature, and it has been an active research area for decades in structural engineering. It is important because it can be used to quickly find preliminary designs for further detailed investigation and design~\cite{stolpe2016truss}

Bilevel optimisation~\cite{sinha2017review} has been an effective approach for tackling truss optimisation problems, while dealing with different design variables and considering interactions among them~\cite{Islam2017,ahrari2020}.

Topology optimisation trusses is a combinatorial and multi-modal problem that for bilevel optimisation of trusses, we consider the topology search space in the upper level and size and shape optimisation of the truss in the lower level. 

\subsection{Related Work}
Truss optimisation problems are subject to structural constraints such as stability, failure criteria and design codes by practice and manufacturing specifications. The constraints can conflict with objectives which makes the problem more challenging.

For decades, several numerical methods have been applied to different truss optimisation problems. Conventional methods such as gradient-based showed limited efficiency in dealing with structural constraints and handling the discreteness and discontinuities in truss optimisation problems which led to trapping in local optima~\cite{DEB2001447}. The inadequacy of classical optimisation methods led to the development of population-based algorithms and metaheuristics in truss optimisation~\cite{kicinger2005evolutionary}.

Two-stage approaches for topology, size and shape optimisation in the literature considers these design variables are linearly separable and initially find optimal topology considering fixed equal size for all active bars. Next, it determines the optimal sizing for the obtained topology. The assumption of linear separability can lead to missing out good solutions and may result in infeasible solutions with respect to real practice~\cite{DEB2001447}. 

Other approaches, namely single-stage approaches consider topology, size and shape design variables simultaneously and consequently expand the search space considering both design variables. The search is guided towards finding one ultimate single solution and still may ignore interactions among different variables.

Bilevel optimisation is a more effective approach than previous two approaches to deal with truss optimisation problems because it can model the interaction among different aspects of the problem more explicitly. In the bilevel formulation, the upper level optimisation problem determines the truss configuration, such as topology, where the lower level optimises bars' sizing.

Islam et al.~\cite{Islam2017} adopted a bilevel representation for the truss optimisation problem, where the weight of the truss was the main optimisation objective for both upper and lower levels. In the upper level, a modified binary Particle Swarm Optimisation (PSO) with niching capability was used to enhance its population diversity while still maintaining some level of exploitation. The niching technique was based on a speciation scheme that applies a niching radius to subdivide the swarm, in an effort to locate multiple high-quality solutions. For the lower level, a standard continuous PSO was employed to supply good sizing solutions to the upper level.

However, using standard PSO for such constrained engineering problems can lead to trapping in local optima~\cite{he2004improved}. It has been showed that truss optimisation should incorporate domain-specific information of the problem instead of only considering pure optimisation \cite{ahrari2016improved}.

It has been observed that there exist multiple distinct topologies with almost equal overall weight in the truss optimisation search space~\cite{DEB2001447,Islam2017,li2016seeking}. Therefore finding multiple equally good truss designs with respect to the topology, size and shape can enable practitioners to choose according to their preferences.

Niching methods~\cite{li2016seeking} and novelty search~\cite{lehman2008exploiting} are capable of finding multiple optima in multi-modal search space. Niching techniques employ rules to keep diversity in the population by primarily focusing on optimising a fitness function.
However, novelty search drives the search towards different candidate solutions from the ones previously have been encountered. Novelty search was proposed in~\cite{lehman2008exploiting} enhances the ability of exploration in population-based algorithms, and keeps the diversity by looking for more novel solutions considering their similarity to other previously visited solutions. Novelty search aims to improve diversity with respect to the behavioural space instead of exclusively considering objective function value~\cite{martinez2019hybridizing,lehman2011abandoning}.

\subsection{Our Contribution}

In this paper, we consider the bilevel optimisation of trusses with a primary focus on the upper level as a combinatorial optimisation problem. We propose two approaches considering the upper level search space in the truss test problems.
We introduce an exact enumeration approach for rigorous analysis of the upper level search space for small test problems. Exact enumeration iterates over all possible upper level topologies, and we apply the lower level optimisation to every feasible upper level design.
Using exact enumeration enables us to remove randomness in the upper level and better characterise its search space and report on the quality of potential designs. 

For larger problems, we propose a novelty-driven binary particle swarm optimisation for bilevel truss optimisation. We aim to discover new designs at the upper level by maximising novelty, and we apply the lower level optimisation to obtained novel solutions.
Using novelty search can guide the search in the upper level towards unseen topologies instead of using only the overall weight to explore the search space. Therefore we set different objective functions for the upper and lower levels.
Our proposed novelty search driven binary PSO for bilevel optimisation of trusses consists of a modified binary PSO in the upper level to deal with the exploration of topology search space and maximise novelty. The upper level of truss optimisation is subject to primary essential constraints. 

We employ a repair mechanism to fix infeasible produced solutions. 
As lower level sizing in truss optimisation is challenging due to the nature of the problem, we use a reliable evolutionary optimiser that incorporates domain-based knowledge to determine the size of bars.

We carry out the proposed approaches to truss optimisation test problems. Our experimental investigation shows that our approaches can outperform the current state of art methods and achieve multiple high-quality truss designs. The source code is available at \url{https://github.com/hiraaaad/BinaryNoveltyPSOTruss}.

The rest of the paper is organised as follows. In the next section, we state the bilevel truss optimisation problem, explain the lower level optimiser and give background on standard and binary PSO. Afterwards, we introduce the exact enumeration and propose the bilevel novelty search framework including the upper level repair operator. We carry out experimental investigations and report on the quality of obtained solutions for different truss test problems. We finish with some concluding remarks and some suggestions for future work.

\section{Bilevel truss optimisation problem}
In this section, we define the bilevel truss problem according to~\cite{Islam2017}. Next, we explain the lower level optimiser, and afterwards, we provide some preliminaries on binary PSO used in~\cite{Islam2017} and this study. Bilevel truss problem embeds lower level size optimisation problem into an upper level topology optimisation problem.

\subsection{Problem Definition}
Bilevel truss optimisation problem embeds an upper level topology optimisation problem into a lower level size optimisation problem. The bilevel optimisation problem can be stated as,
\begin{align*}
	\text{find} &\quad \vec{x},\vec{y}\text{,}  \quad \vec{x}\in \{0,1\}^m\\
	\text{optimise} &\quad F(\vec{x},\vec{y}) \\
	\text{subject to} & \quad G_1(\vec{x}),\, G_2(\vec{x}),\, G_3(\vec{x}) \\
	\text{where} &\quad  G_1(\vec{x}) = \text{True} \iff  \text{Necessary nodes are active in truss} \\
	            & \quad G_2(\vec{x}) = \text{True} \iff  \text{Truss is externally stable} \\
                 & \quad G_3(\vec{x}) = \vec{y} \in \text{argmin}\{W(\vec{x},\vec{y}),\, g_j(\vec{x},\vec{y}) \leq 0, \, j = 1,\,2,\,3\}
\end{align*}
where $\vec{x}$ refers to the binary topology variable in the upper level where it shows if a truss bar is active (1) or excluded (0). $m$ shows the length of upper level topology design variables. For instance, in 25-bar truss $m$ is 8 due to symmetry. For the same problem, we can show the upper bound of topology as the ground structure where all bars are active as $\vec{x}=[11111111]$.

$\vec{y}$ denotes the design variable in the lower level optimisation problem including size and shape with respect to the test problem. The size variables of $\vec{y}$ should be selected from an available size set ($S$). $F(\vec{x},\vec{y})$ shows the objective function considered in the upper level such as weight minimisation used in~\cite{Islam2017} or maximising novelty in this study.

\begin{figure*}
	\centering
 	\includegraphics[width=0.9\textwidth]{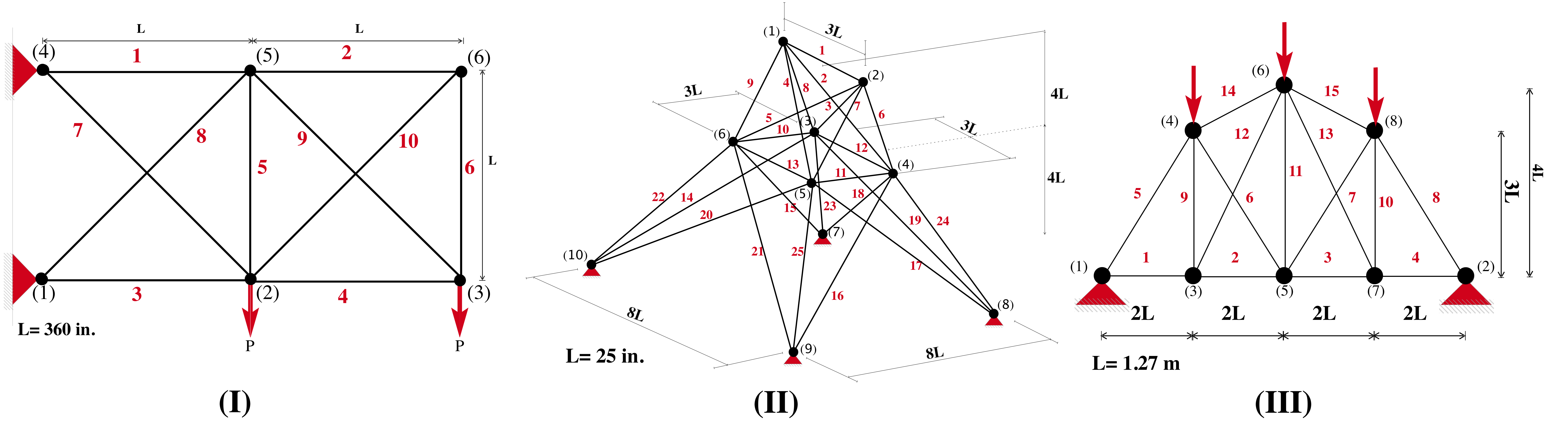}
	\caption{Ground structure of 10-bar truss (I), 25 bar truss (II) and 15-bar truss (III).}
	\label{figure.TrioGS}
\end{figure*}

Solutions in the upper level should satisfy the topology constraints for feasibility. $G_1$ enforces that the design should have active nodes that support the truss and carry the external load, because they are necessary elements in the design space's predefined boundary conditions. For example for 10-bar truss (depicted in Figure~\ref{figure.TrioGS} (I)) nodes 1 and 4 are support nodes and nodes 2 and 3 are carrying external loads. Therefore these four nodes are necessary nodes in the design space. 
$G_2$ states the external stability of a truss. The external stability satisfaction criteria are fully detailed in Section~\ref{subsec:repair}. Feasible topology solutions should meet $G_1$ and $G_2$. In this case, the lower level optimiser aims to find the optimum $\vec{y}$ to minimise the overall weight of truss ($W$) which is the summation of the weight of 
all included bars ($\hat{m}$) in the truss.
\[W(\vec{x},\vec{y})=\rho\sum_{i=1}^{\hat{m}}x_{i}y_{i}l_{i}\]
where $\rho$ and $l$ show the density of the material used in the truss (such as aluminium or steel) and length of a bar with respect to its end points in the design space, respectively.
upper level external stability satisfaction is necessary but not sufficient. Therefore, in the lower level, the internal stability should be checked through lower level function evaluation. 

If $(\vec{x},\vec{y})$ meets the internal stable condition, extra constraints should be satisfied. These constraints state that the computed stress in bars ($\sigma_i$, $i \in\{1,2,..,\hat{m}\}$) and displacement of truss nodes ($\delta_k$, $k=\{1,2,..,n\}$) after applying the external loads should not exceed their problem-dependent allowable values $\sigma_i^{max}, i \in\{1,2,..,\hat{m}\}$, and $\delta_k^{max}, k \in\{1,2,..,n\}$, respectively.

The lower level constraints $g_j$, $j = 1, 2, 3$, used as part of $G_3(\vec{x})$ are therefore defined as follows.
\begin{align*}
g_1(\vec{x},\vec{y}) &= \text{True} \iff  \:\text{Truss is internally stable} \\
g_2(\vec{x},\vec{y}) &\leq 0 \iff g_{2,i}(\vec{x},\vec{y}) = \sigma_i - \sigma_i^{max} \leq 0 			 \quad \forall i \in\{1,2,..,\hat{m}\} \\
g_3(\vec{x},\vec{y}) &\leq 0 \iff g_{3,k}(\vec{x},\vec{y}) = \delta_k - \delta_k^{max} \leq 0 			 \quad \forall k \in \{1,2,..,n\}
\end{align*}
\subsection{Lower Level Optimisation}
We use a domain knowledge-based evolutionary optimiser for lower level optimisation~\cite{ahrari2016improved} to determine the optimum layout. The \textit{loweroptimiser} is a variant of Covariance Matrix Adaptation Evolution Strategy algorithm (CMA-ES) that is customised to be problem-specific. 
\textit{loweroptimiser} follows the main principles of evolutionary strategies to evolve the solutions. However, it uses specific operators to adjust solutions with respect to the allowable stress and displacement in the truss. \textit{loweroptimiser} uses a probabilistic scheme to round the values to the discrete set to avoid biasing the search towards sub-optimal solutions. 

\textit{loweroptimiser} employs a mapping strategy with respect to the response after performing finite element analysis to adjust the sizing of a violating bar by multiplying its current size with a factor that depends on the amount of violation. 
Another operator is a resizing strategy for producing new individuals near boundary designs of the problem and the problem-dependent constraints. For brevity, we refer the reader to~\cite{ahrari2016improved,ahrari2020} for detailed explanations on different components of the \textit{loweroptimiser}.

\subsection{Particle Swarm Optimisation (PSO)}
PSO is a population-based algorithm evolving a swarm where it contains \textit{particles} as candidate solutions to a problem~\cite{kennedy1995particle}. Each particle has three vectors at the time ($t$) of evolution : position ($\vec{z_t}$), velocity ($\vec{v_t}$) and personal best where it keeps the best position is has been evolved to as ($\vec{p_t}$). PSO updates particle positions based on the velocity for each component ($i$) as follows.
\begin{align*}
    \vec{v}_{t+1} &= \omega \vec{v}_t + c_1 r_1 \times (\vec{p}_t - \vec{z}_t) + c_2 r_2 \times (\vec{p}_g - \vec{z}_t)\\
    \vec{z}_{t+1} &= \vec{z}_t + \vec{v}_{t+1}
\end{align*}
where $\vec{p_g}$ is the global best position in the swarm. $\omega$ is the inertia factor to control the impact of current velocity in velocity updating. $r_1$ and $r_2$ are vectors with size of a particle containing random values from a uniform distribution in the range [0, 1]. $c_1$ and $c_2$ are cognitive and social factors to attract particles toward their personal best and global best. 

\subsection{Binary PSO}
PSO is typically used as a continuous optimisation algorithm. Therefore, to use it for binary search spaces, we need to employ a transfer function, such as Sigmoid transfer function, to map a continuous search space into a binary search space. In this study, we use global topology for the PSO and employ a time-varying transfer function~\cite{Islam2017} to balance between exploration and exploitation. Velocities of all particles are updated according to the velocity update equation. The following equation determines the probabilities for flipping the position vector elements ($i$).
\begin{equation*}
	z_i^t = \begin{cases}
	1 & \text{if $rand() \geq TV(v_i^t,\varphi)$} \\
	0 & \text{otherwise,}
	\end{cases}
\end{equation*}
where $rand()$ denotes a random value in a uniform distribution in the range [0, 1] and, $TV$ is given as~\cite{Islam2017},
\begin{equation*}
	TV(v_i^t,\phi) = \frac{1}{1+e^{{-\frac{v_i^t}{\varphi}}}}
\end{equation*}
$\phi$ is the control parameter to balance exploration and exploitation in the course of evolution where it linearly decreases from 5.0 to 1.0 in this study according to~\cite{Islam2017}.

\section{Exact Enumeration} \label{sec:exact}
We apply exact enumeration to the truss problems where the upper level dimension of the search space is small ($m\leq12$). Exact enumeration enables us to enumerate over all possible combinations of binary strings in the search space in the upper level, where each represents a topology design. Therefore we can remove the randomness for these problems from the upper level and investigate its search space rigorously.
Algorithm~\ref{alg.exact} shows the pseudocode of our exact enumeration. This algorithm takes the upper level dimension $m$ as the input and iterates over all possible binary string combinations of $m$-bits. 
\begin{algorithm}[t]
\caption{Exact Enumeration}\label{alg.exact}
\DontPrintSemicolon
% \KwIn{$m:$ \Comment*[r]{length of topology variable}}
\For{$i = 1;\ i \leq 2^{m};\ i = i + 1$}
{compute $\vec{x_i}$ \Comment*[r]{generate the bit string} 
\If{$\vec{x}_i$ is feasible}{%\Comment*[r]{Binary topology design variable}\\
$\vec{y}_i$ = \textit{loweroptimiser}($\vec{x}_i$) \\
}
Store $W(\vec{x_i},\vec{y_i})$ 
}
\end{algorithm}

If a binary string satisfies the upper level's feasibility criteria $G_1$ and $G_2$, it will be sent to the lower level. \textit{loweroptimiser} finds the optimum vector for size (and shape), and we store its overall weight.

\section{Bilevel Novelty Search}
In this section we first introduce the components of our bilevel method, and then we combine these components and introduce the framework of proposed novelty PSO for bilevel truss optimisation.

\subsection{Novelty-Driven PSO }
Novelty-Driven PSO (NdPSO) is a variant of PSO employing novelty search to drive particles toward novel solutions that are different from previously encountered ones~\cite{NdPSO}. 
The main idea is to explore the search space by ignoring objective-based fitness functions and reward novel individuals. 
NdPSO uses the score of novelty to evaluate the performance of particles. For this purpose, it maintains an archive of past visited solutions to avoid repeatedly cycling through the same series of behaviours. 

NdPSO evaluates the novelty of particles by computing the average distance of a behaviour to its k-nearest neighbours in the archive as follows:
\[\text{nov}(x) = \frac{1}{k}\sum_{i=0}^{k}\textit{dist}(x,\mu_i).\] where $\mu_i$ is the $i$\textsuperscript{th} nearest neighbour of $x$ and \textit{dist} is the Hamming distance metric. Novelty score ensures that individuals in less dense areas with respect to the archive get higher novelty scores.

NdPSO employs core principles of PSO and mainly replaces the objective function with novelty evaluation. Note that personal best and global best value in NdPSO show a dynamic behaviour. For more details on NdPSO, we refer the reader to~\cite{NdPSO}. We use NdPSO in the upper level of truss optimisation to discover novel topology designs.
%%%%%%%%%%%%%%%%%%%%%%%
\subsection{Repair Mechanism in the Upper Level} \label{subsec:repair}
Topology designs in the upper level are feasible if they meet $G_1(\vec{x})$ and $G_2(\vec{x})$.

The following conditions should be satisfied for $G_2(\vec{x})$~\cite{ahrari2016improved}:
\begin{itemize}
    \item The degree of freedom (DoF) in a truss should be non-positive~\cite{DEB2001447}.
    \item The summation of the number of members and restrain forces on a node must be equal or greater than the truss dimension.
    \item The summation of the number of members and restraint forces on a non-carrying node must be greater than the truss dimension.
\end{itemize}
As stated in~\cite{DEB2001447}, necessary node constraints are more important than the DoF constraint. To deal with infeasible topologies, we use the (1+1)-EA~\cite{DROSTE2002} with the following comparator to repair solutions:
\begin{align*}
x \succeq y :=  &\left(\alpha(x) \leq \alpha(y)\right)\vee \\
    &\left(\alpha(x)=\alpha(y)  \wedge \beta(x)\leq \beta(y)\right) \vee \\
    &\left(\alpha(x)=\alpha(y) \wedge \beta(x)=\beta(y) \wedge \theta(x) \leq \theta(y)\right)
\end{align*}
where $\alpha$ is the violation degree of active necessary nodes, $\beta$ is the violation degree of truss DoF and $\theta$ is the violation degree of second and third criteria in external stability. (1+1)-EA is a simple evolutionary algorithm where it produces an offspring by mutation and the offspring replaces the parent if the offspring is better with respect to the objective function mentioned above.
\subsection{Bilevel Novelty-Driven Binary PSO Framework}
\begin{algorithm}[t]
\caption{Novelty-Driven Binary PSO for Bilevel Truss Optimisation}\label{alg.NS}
\DontPrintSemicolon
Randomly generate the initial population of Binary PSO \\
Repair the initial population into feasible topologies \\
Set the velocity of particles in population \\
Evaluate the novelty score for each particle\\
$\vec{y}_i$ = \textit{loweroptimiser}($\vec{x}_i$) \\
Store $W(\vec{x}_i,\vec{y}_i)$ \\
Update $p_t$ and $p_g$ \\
Update the archive \\
\Repeat{termination criterion is met}{
\For{i=1 to population size}
{
Update position of particle \\
Update velocity of particle \\
Repair the particle into feasible upper level solution \\
Evaluate novelty score of the particle \\
$\vec{y}_i$ = \textit{loweroptimiser}($\vec{x}_i$) \\
Store $W(\vec{x_i},\vec{y_i})$ \\
Update $p_t^i$ and $p_g$ according to novelty score \\
Update the archive \\
}
}
\end{algorithm}
Our proposed approach works as follows (see Algorithm~\ref{alg.NS}). 
Initially, the binary PSO generates a random population of binary strings. Next, the repair mechanism performs on the population to ensure the feasibility of particles. The particles' velocities are drawn randomly from $[-\upsilon,\upsilon]$. Then, the novelty score is computed for particles with respect to the archive. 
Because all particles are feasible, \textit{loweroptimiser} computes the corresponding optimal size (and shape) for the upper level topology. With this, we update the archive with the current population. Next, the position and velocity of particles are updated, and the above process repeats till the termination criterion is met. 
\section{Experimental investigations}
In this study, we use multiple truss test problems with discrete sizing from the literature~\cite{Li2009,Degertekin2019,ahrari2016improved}. We investigate them in ascending order of length of topology design variable.

We split the problems into small and large instances. We apply exact enumeration to small problems where their topology search space is tractable ($m\leq12$). To show its outcome, we set the ground structure of the problem as an upper bound reference. We sort the other designs with respect to their Hamming distance with this reference design (denoted by $d_H$). We report on the quality of solutions using the median of 30 independent runs in the lower level. 

For large instances, we apply the proposed bilevel novelty search PSO and report on the quality of top best-found solutions. We investigate the obtained designs and identification of redundant bars and nodes in the design space.
To setup our algorithms, we use the following parameters. For lower level optimisation, we use the parameter settings in~\cite{ahrari2016improved,ahrari2020}. For the upper level optimisation, the swarm consisted of 30 particles, $\upsilon=6$~\cite{Islam2017}, $c_1=c_2=1.0$, $\omega$ linearly decreases from 0.9 to 0.4, and the maximum number of iterations is set to 300. We use $k=3$ nearest neighbours to calculate novelty at the upper level.% We split the problems into small and mid scale instances.
\subsection{25-bar truss}
Figure~\ref{figure.TrioGS} (II) shows the ground structure of 25-bar truss which is a spatial truss for size and topology optimisation \cite{Degertekin2019}. This problem splits into two cases with respect to different load cases. The truss undergoes a single and multiple external loads for the first and second cases, respectively. The sizing of bars should be selected from different discrete sets for each case \cite{Degertekin2019}. 

\begin{figure}[t]
	\centering
	\includegraphics[width=0.75\textwidth]{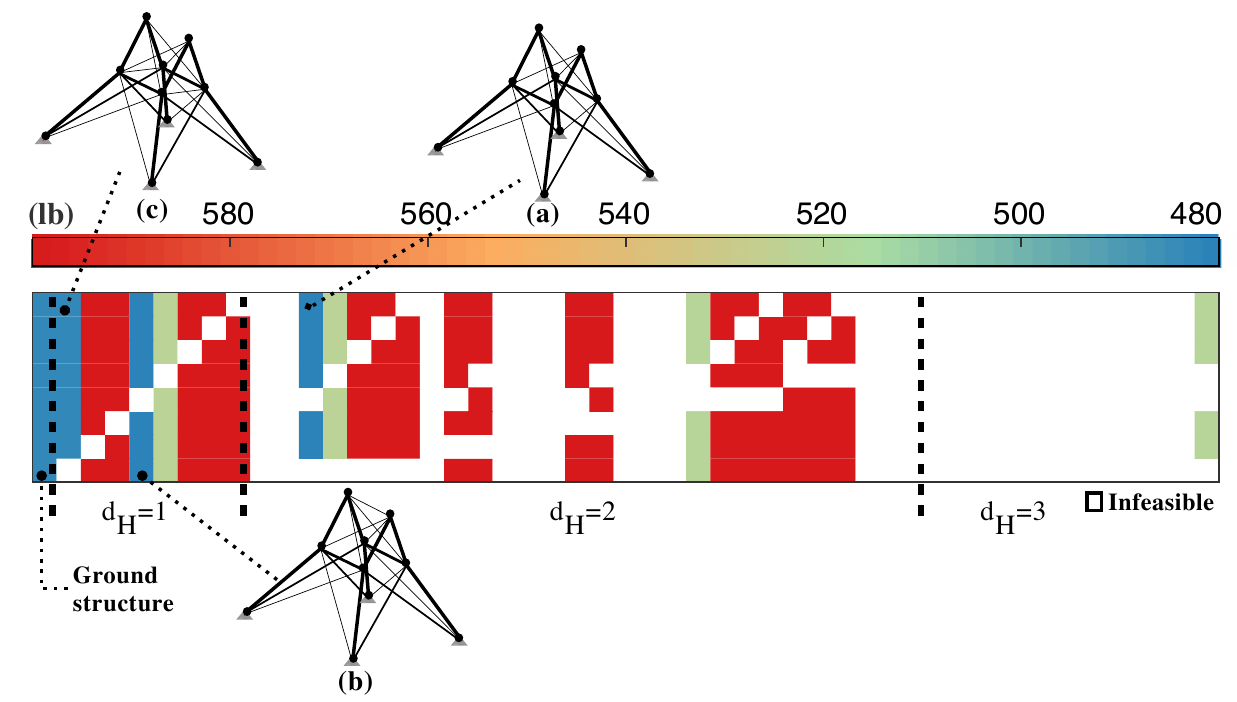}
	\caption{Exact enumeration on 25-bar truss case 1 (right side truncated). $d_H$ denotes the hamming distance with the upper bound reference.}
	\label{figure.25SS1}
\end{figure}

\begin{figure}[t]
	\centering
	\includegraphics[width=0.75\textwidth]{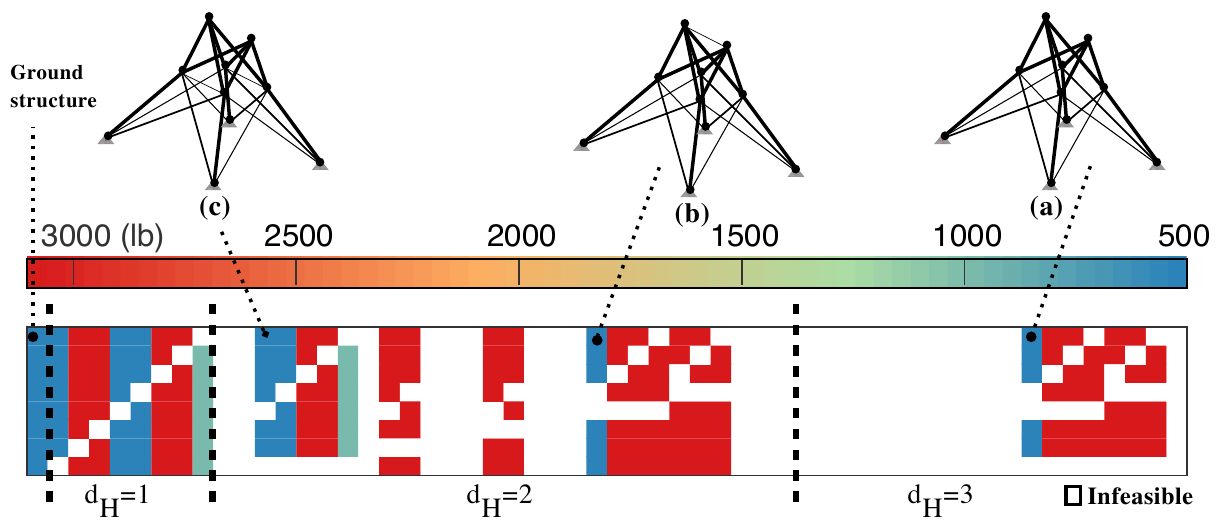}
	\caption{Exact enumeration on 25-bar truss case 2 (right side truncated).$d_H$ denotes the hamming distance with the upper bound reference.}
	\label{figure.25SS2}
\end{figure}

This is a symmetric truss where 25 truss bars are grouped into eight groups of bars. Therefore, there are 256 possible topologies at the upper level. 
Figures~\ref{figure.25SS1} and~\ref{figure.25SS2} show the outcome of exact enumeration on this test problem where we observed that up to first 50 and 55 designs with respect to $d_H$ are feasible and the rest of the search space is infeasible. We can see that most of the feasible designs are in the upper bound vicinity where $d_H \leq 3$. We can also observe that the high-quality solutions exist in the region where $d_H \leq 2$ and  $d_H \leq 3$ for cases 1 and 2, respectively.

\begin{table*}[t]
\centering
\caption{Comparison of optimised designs for 25-bar truss.}
\label{table.25barmerge}
\begin{adjustbox}{max width=\textwidth}
%\centering
\setlength{\tabcolsep}{6pt}
\begin{tabular}{lllllllll}
\toprule
 &
   &
  \cite{Rajeev1992} &
  \multicolumn{3}{c}{\cite{HoHuu2016}, \cite{Cheng2016}, \cite{Degertekin2019}} &
  \multicolumn{3}{c}{This Study} \\
  \midrule
Case 1  &  &  & \multicolumn{3}{l}{} & (a) & (b) & (c) \\
\cmidrule{7-9}
%   &
  \textbf{Best weight (lb)} &&
  546.01 &
  \multicolumn{3}{l}{484.85} &
  482.6 &
  483.35 &
  484.3 \\ 
  \midrule
  & & \cite{Lee2005} & \multicolumn{3}{l}{\cite{Li2009}, \cite{Degertekin2019}} &
  \multicolumn{3}{c}{This Study} \\
\midrule
Case 2 &  &  & \multicolumn{3}{l}{} & (a) & (b) & (c) \\
\cmidrule{7-9}
%   &
  \textbf{Best weight (lb)} &&
  556.43 &
  \multicolumn{3}{l}{551.14} &
  546.97 &
  547.81 &
  548.64 \\
\bottomrule
\end{tabular}

\end{adjustbox}
\end{table*}

Table~\ref{table.25barmerge} shows our findings for both cases.
\ignore{For the first case, we can see that designs (b) and (c) both have identified two solutions with $d_H = 1$. However, Design (b) identifies two bars $A_{10}-A_{11}$ as redundant, but design (c) only identifies $A_1$ as redundant. For the second case, both designs (b) and (c) commonly identify $A_{10}-A_{11}$ as redundant.
Design (a) in both cases represents the best-found solution where it combines the identified redundant bars in corresponding design (b) and (c) in each case. It also adjusts the sizes of bars to meet truss constraints leading to a lightweight solution.}  

% \begin{table}[] 
% \caption{Comparison of optimised designs for 25-bar truss case 1.}
% \label{table.25bar}
% \begin{adjustbox}{max width=\columnwidth}
% \centering
% \input{table_25.tex}
% \end{adjustbox}
% \end{table}
% case 2
\subsection{10-bar truss}
10-bar truss is a well-known size and topology optimisation problem which its ground structure is depicted in Figure~\ref{figure.TrioGS} (I). It undergoes single load and the sizing of bars should be selected from a discrete set where There are 10 bars in the topology design, which results in 1024 possible upper level topologies. See \cite{Li2009} for details on the simulation.

\begin{figure}[t]
	\centering
	\includegraphics[width=0.9\textwidth]{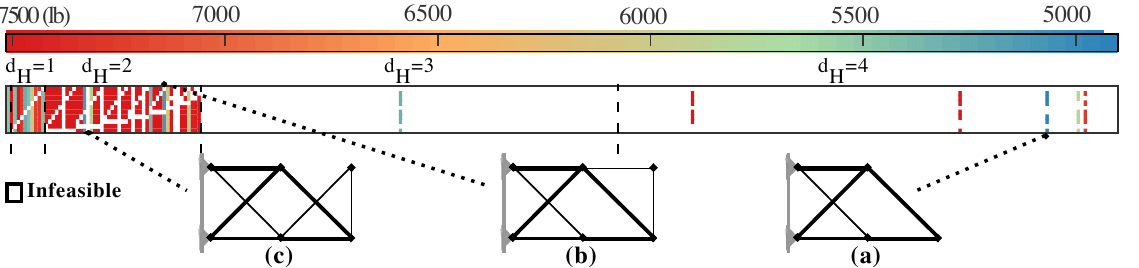}
	\caption{Exact enumeration on 10-bar truss. $d_H$ denotes the hamming distance with the upper bound reference.}
	\label{figure.10SS}
\end{figure}

Figure~\ref{figure.10SS} shows the outcome of exact enumeration and the designs are ordered according to their $d_H$. We only show the first 320 sorted designs because the rest of the designs are infeasible. This figure also shows the top obtained designs for this problem.
We can see feasible designs are where $d_H \leq 4$ and the best-found design's $d_H$ is 4. Table~\ref{table.10bar} shows our findings, and we can see that designs (b) and (c) with $d_H=2$, both identify two bars as redundant and incorporate all six nodes in their designs. 
\begin{table*}[t]
\caption{Comparison of optimised designs for 10-bar truss.}
\label{table.10bar}
\begin{adjustbox}{max width=\columnwidth}
\centering
\setlength{\tabcolsep}{6pt}
\begin{tabular}{@{}lllllllllll@{}}
\toprule
% A : in\textsuperscript{2}
& \cite{Li2009}  & \multicolumn{3}{l}{\cite{Cheng2016}, \cite{HoHuu2016}, \cite{Degertekin2019}} & \cite{Fenton2014} & \cite{Khayyam2020} & \multicolumn{3}{c}{This Study} \\
\midrule
&                &  \multicolumn{3}{l}{} & & & (a)       & (b)       & (c)       \\
\cmidrule{8-10}
\textbf{\shortstack{Best weight (lb)}} & 5531.98 & \multicolumn{3}{l}{5490.74 }& 5056.88       & 4980.10     & 4965.70   & 5107.50   & 5131.70  \\
\bottomrule
\end{tabular}
\end{adjustbox}
\end{table*}
However, design (a) which is the best-found design identifies four bars ($A_2$, $A_5-A_6$ and $A_{10}$) as redundant and eliminates node 6. \ignore{Note that node 6 is not a necessary node and can be eliminated from the design space.} We can also see that other state of art methods, including~\cite{Fenton2014} and~\cite{Khayyam2020} also identified the same topology as the optimal topology. However, our approach can obtain a solution with a lower weight due to the efficient lower lover optimiser.

\subsection{52-bar truss}
\begin{figure}[t]
	\centering
	\includegraphics[width=0.8\textwidth]{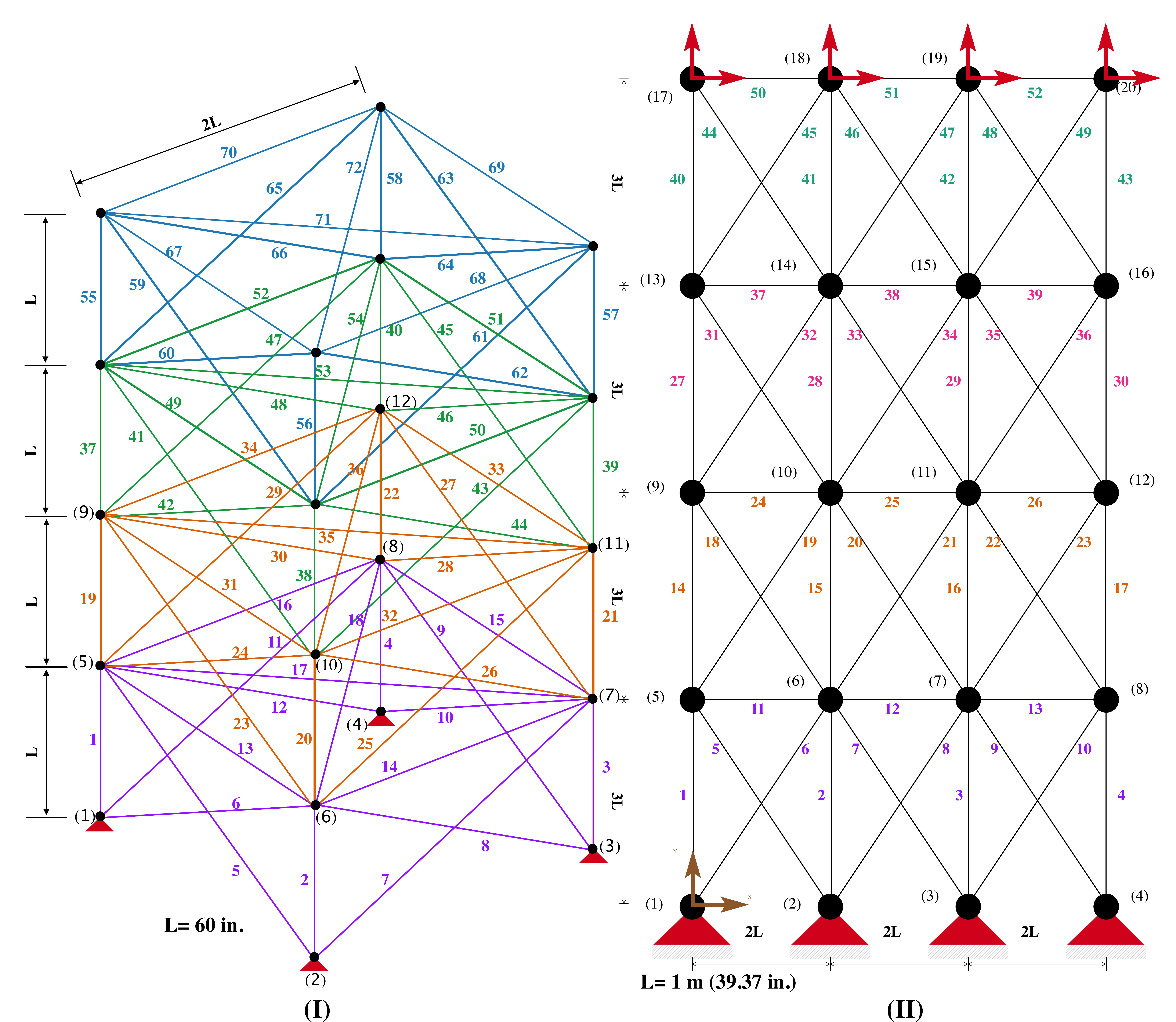}
	\caption{Ground structure of 52-bar truss (I) and 72-bar truss (II).}
	\label{figure.5272GS}
\end{figure}

\begin{figure}[t]
	\centering
 	\includegraphics[width=0.75\textwidth]{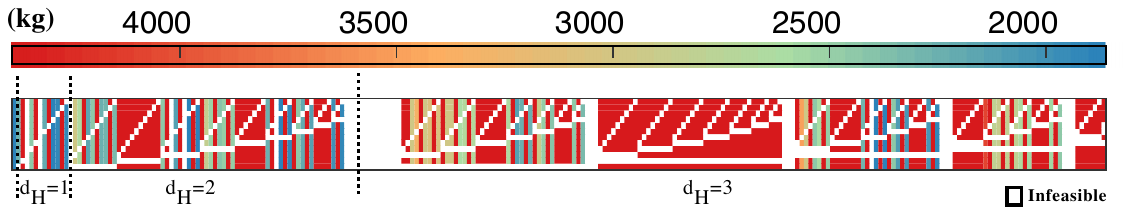}
	\caption{Exact enumeration on 52-bar truss (right side truncated).}
	\label{figure.52SS}
\end{figure}

\begin{figure}[t]
	\centering
	\includegraphics[width=0.7\textwidth]{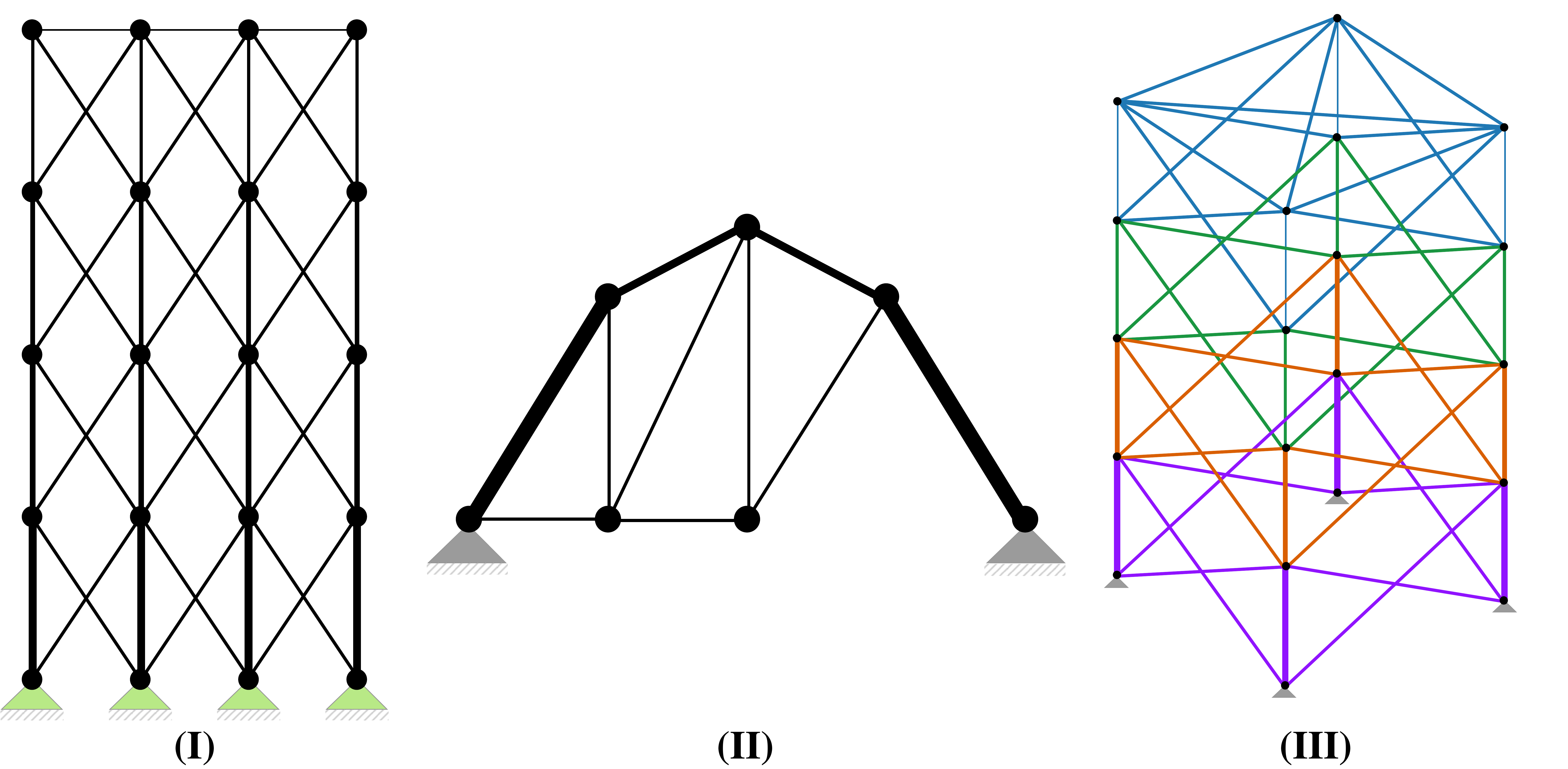}
	\caption{Best-found designs for 52-bar (I), 15-bar truss (II) and 72-bar (III).}
	\label{figure.topGS}
\end{figure}

This is a size and topology optimisation problem where 52-bar truss are grouped into 12 bar groups resulting in 4096 possible designs. The truss undergoes three load cases and the sizing of bars should be selected from a discrete set \cite{Wu1995}. Figure~\ref{figure.5272GS} (I) shows the ground structure of 52-bar truss, which is composed of 12 bars resulting in 4096 possible designs.
Figure~\ref{figure.52SS} shows the outcome of exact enumeration where we only show the first 250 designs sorted according to their $d_H$. For clear presentation, Out of all sorted combinations, 1900 designs are feasible, and the rest of the search space is infeasible. We also observed that feasible designs are located where $d_H \leq 6$.
Table~\ref{table.52bar} shows our findings and, we can see that the top three designs for this problem identify one to three group of bars redundant, respectively.
\begin{table*}[t]
\caption{Comparison of optimised designs for 52-bar truss.}
\label{table.52bar}
\begin{adjustbox}{max width=\textwidth}
\centering
\setlength{\tabcolsep}{3pt}
\begin{tabular}{lllllllllll}
\toprule
% A : mm\textsuperscript{2}
& \cite{Wu1995}   & \cite{Lee2005}  & \cite{Li2009}   & \cite{Kaveh2009} & 
\multicolumn{3}{l}{\cite{HoHuu2016}, \cite{Cheng2016}, \cite{Degertekin2019}} & \multicolumn{3}{c}{This Study} \\
\midrule
&&&&&&& & (a) & (b) & (c) \\
\cmidrule{9-11}
\textbf{Best weight (kg)} & 1970.14 & 1906.76 & 1905.5 & 1904.830  & \multicolumn{3}{l}{1902.610} & 1862  & 1880.3  & 1869.7 \\
\bottomrule
\end{tabular}
\end{adjustbox}
\end{table*}
Design (a) depicted in Figure~\ref{figure.topGS} (I) shows the best-found design that eliminates all redundant bars identified in design (b) and (c) and removes $A_{37}-A_{39}$ summing up to nine redundant bars.

\subsection{15-bar truss}
This is a size and topology truss optimisation problem which is non-symmetric truss composed of 15 bars and truss undergoes three load cases and the sizing should be selected from a discrete set \cite{Zhang2005}. Figure~\ref{figure.TrioGS} (III) shows the ground structure of 15-bar truss which is a non-symmetric truss composed of 15 bars.

Table~\ref{table.15bar} shows our findings by the proposed bilevel novelty search compared with other methods. We can see that designs (b) and (c) find the same weight, and they are symmetric around the vertical axis with respect to the topology and size of bars. Both designs remove 6 bars from the design space, and symmetrically they eliminate nodes 2 and 4 from the design space, respectively. Design (d) eliminates five bars and node 2 from the design space. 

\begin{figure}[h]
	\centering
\includegraphics[width=0.9\textwidth]{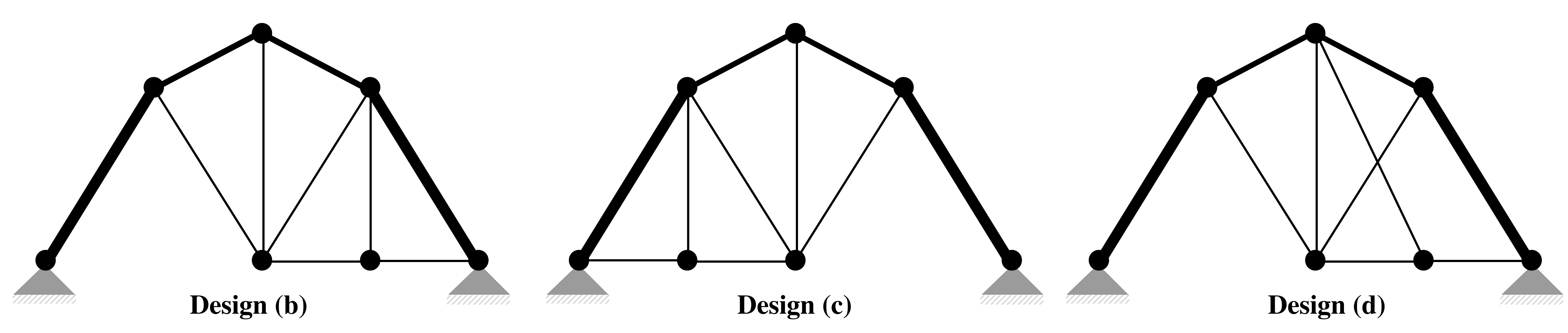}
	\caption{Second to fourth (b-d) best-found designs for 15-bar truss.}
	\label{figure.top15}
\end{figure}

Figure~\ref{figure.top15} shows these three designs.
Design (a) is the best-found design depicted in Figure~\ref{figure.topGS} (II) eliminates five bars in the design space and provides a lighter solution compared with other methods.

\begin{table}[t]
\centering
\caption{Comparison of optimised designs for 15-bar truss.}
\label{table.15bar}
\begin{adjustbox}{max width=\columnwidth}
\setlength{\tabcolsep}{2pt}
\begin{tabular}{lllllll}
\toprule
% A : mm\textsuperscript{2}
& \cite{Zhang2005}& \cite{Cheng2016} & \multicolumn{4}{c}{This Study}        \\
\midrule
          &      &      & (a)          & \multicolumn{2}{c}{(b), (c)}  & (d)\\
\cmidrule{4-7}
\textbf{Best weight (kg)} & 142.12    & 105.74    & 89.899         & \multicolumn{2}{c}{90.223}   & 91.874      \\
\bottomrule
\end{tabular}
% $A_{1}$          & 308.6     & 113.2     & 113.2          & -             & 113.2       & -\\
% $A_{2}$          & 174.9     & 113.2     & 113.2          & -             & 113.2       & - \\
% $A_{3}$          & 338.2     & 113.2     & -              & 113.2         & -           &113.2\\
% $A_{4}$          & 143.2     & 113.2     & -              & 113.2         & -           &113.2\\
% $A_{5}$          & 736.7     & 736.7     & 736.7          & 736.7         & 736.7       &736.7\\
% $A_{6}$          & 185.9     & 113.2     & -              & 113.2         & 113.2       &113.2\\
% $A_{7}$          & 265.9     & 113.2     & 143.2          & 143.2         & 143.2       &143.2\\
% $A_{8}$          & 507.6     & 736.7     & 736.7          & 736.7         & 736.7       &736.7\\
% $A_{9}$          & 143.2     & 113.2     & 113.2          & -             & 113.2       &-\\
% $A_{10}$         & 507.6     & 113.2     & -              & 113.2         & -           &-\\
% $A_{11}$         & 279.1     & 113.2     & 113.2          & 145.9         & 145.9       &145.9\\
% $A_{12}$         & 174.9     & 113.2     & 113.2          & -             & -           &-\\
% $A_{13}$         & 297.1     & 113.2     & -              & -             & -           &113.2\\
% $A_{14}$         & 235.9     & 334.3     & 334.3          & 334.3         & 334.3       &334.3\\
% $A_{15}$         & 265.9     & 334.3     & 334.3          & 334.3         & 334.3       &334.3\\
\end{adjustbox}
\end{table}

\subsection{72-bar truss}
\begin{table*}[t]
\centering
\caption{Comparison of optimised designs for 72-bar truss.}
\label{table.72bar}
\begin{adjustbox}{max width=\columnwidth}
\setlength{\tabcolsep}{4pt}
\begin{tabular}{llllllll}
\toprule
% A : in.\textsuperscript{2}
& \cite{Wu1995} & \cite{Lee2005} & \multicolumn{2}{l}{\cite{Cheng2016}, \cite{Degertekin2019}} & \multicolumn{3}{c}{This Study} \\
\midrule
&&&& & (a) & (b) & (c)\\
\cmidrule{6-8}
\textbf{Best Weight (lb)}                            & 400.66                         & 387.94                          & \multicolumn{2}{c}{385.54}                            & 368.16   & 369.15   & 370.15   \\
\bottomrule
\end{tabular}
\end{adjustbox}
\end{table*}
72-bar truss represents a four storey structure for size and topology optimisation where which is a symmetric truss composed of 72 bars grouped into 16 groups of bars. The truss undergoes two load cases and the sizing of bars should be selected from a set \cite{Wu1995}. 

Table~\ref{table.72bar} shows our findings by the proposed bilevel novelty search compared with other methods. We can see that designs (b) and (c) identify five groups of bars as redundant, including four common groups. This will lead to the elimination of 16 bars from the design space.
Design (a) depicted in Figure \ref{figure.topGS} (III) is the best-found design. It combines the identified redundant bars in designs (b) and (c) and removes 20 bars in total from the design space and achieves a lighter solution. 

\subsection{47-bar truss}
\begin{table*}[t]
\centering
\caption{Comparison of optimised designs for 47-bar truss.}
\label{table.47bar}
\begin{adjustbox}{max width=\columnwidth}
\setlength{\tabcolsep}{2pt}
\begin{tabular}{@{}llllllll@{}}
\toprule
% A : in\textsuperscript{2}
& \cite{hasancebi2001} & \cite{Panagant2018} & \cite{degertekin2018sizing} & \cite{ahrari2016improved} & \multicolumn{3}{c}{This Study} \\
\midrule
&       & &                  &           & (a)       & (b)       & (c)       \\
\cmidrule{6-8}
\textbf{\shortstack{Best weight (lb)}} & 1885.070 & 1871.700 & 1836.462 & 1727.624 & 1724.947               & 1726.044               & 1727.624 \\
\bottomrule
\end{tabular}
\end{adjustbox}
\end{table*}
This problem represents a transmission tower with symmetric truss bars grouped into 27 groups, and it considers size and shape optimisation in the lower level. See \cite{hasancebi2001} for structural constraints and technical information to simulate the problem. Table \ref{table.47bar} shows our findings in comparison with state of the art. We can see that our method obtained three different designs with respect to the topology. Design (a) incorporates 23 topology bars, and design (b) and (c) both incorporate 21 groups of bars and the main difference is including member groups of 23 and 26, respectively.
\subsection{200-bar truss}
\begin{table*}
\centering
\caption{Comparison of optimised designs for 200-bar truss where $\dagger$ denotes the reported solution is infeasible.}
\label{table.200bar}
\begin{adjustbox}{max width=\columnwidth}
\setlength{\tabcolsep}{4pt}
\begin{tabular}{@{}llllllll@{}}
\toprule
% A : in\textsuperscript{2}  
& \cite{Cheng2016} & \cite{HoHuu2016} & \cite{kaveh2009particle} & \cite{Degertekin2019}  & \multicolumn{3}{c}{This Study} \\
\midrule
   &           &           &  & & (a) & (b) & (c) \\
\cmidrule{6-8}
\textbf{\shortstack{Best weight (lb)}} & 27163.59 & 27858.50 & 25156.50$^\dagger$ & 27282.54& 27144.0 & 27575.0 & 27744.0 \\
% \textbf{\shortstack{Constraint violation}} &           &           &  &            &            &                        &                       \\
\bottomrule
\end{tabular}

\end{adjustbox}
\end{table*}
This size and topology optimisation problem includes 200-bars grouped into 29 bar groups. The truss undergoes three loading conditions and is subject to no displacement constraint. See \cite{kaveh2009particle} for details on simulation. Table \ref{table.200bar} shows our findings and compared with reported weights in state of the art. The best-reported weight is 25156.5 lb, but this is an infeasible design because it violates the stress constraint by about 8\%. We can see that the best design obtained by our method (design (a)) outperforms other designs. Design (a) and (c) both include 24 group bars but with different topologies; however, Design (b) incorporates all group members as 29.
\subsection{224-bar truss}
\begin{table*}[tbp!]
\centering
\caption{Comparison of optimised designs for 224-bar truss.}
\label{table.224bar}
\begin{adjustbox}{max width=\columnwidth}
\setlength{\tabcolsep}{2pt}
\begin{tabular}{@{}lllllll@{}}
\toprule
% A : in\textsuperscript{2} 
& \cite{hasancebi2001} &
  \cite{HASANCEBI2002} & \cite{ahrari2016improved}  & \multicolumn{3}{c}{This Study} \\
\midrule
   &     & &  & (a) & (b) & (c) \\
\cmidrule{5-7}
\textbf{\shortstack{Best weight (lb)}} & 5547.500 & 4587.290 & 3079.446 & 3063.866 & 3079.446 & 3102.079 \\
\bottomrule
\end{tabular}

\end{adjustbox}
\end{table*}
This problem represents a pyramid truss where the truss bars are grouped into 32 groups and considers size, shape and topology optimisation subject to complex design specifications. See \cite{hasancebi2001} for details on the simulation. Table \ref{table.224bar} lists obtained designs by our method compared with state of the art. Design (a) outperforms other designs, and design (b) is as alike as the optimum design obtained in \cite{ahrari2016improved}. 
\subsection{68-bar truss}
\begin{table*}[tbp!]
\centering
\caption{Comparison of optimised designs for 68-bar truss.}
\label{table.68bar}
\begin{adjustbox}{max width=\columnwidth}
\setlength{\tabcolsep}{4pt}
\begin{tabular}{@{}llllll@{}}
\toprule
% A : in\textsuperscript{2}
& \cite{Panagant2018} & \cite{ahrari2016improved} & \multicolumn{3}{c}{This Study} \\
\midrule
& &          & (a)       & (b)       & (c)       \\
\cmidrule{4-6}
\textbf{\shortstack{Best weight (lb)}} & 1385.800     & 1166.062   & 1166.062               & 1167.528               & 1169.039 \\
\bottomrule
\end{tabular}
\end{adjustbox}
\end{table*}
This size, shape and topology optimisation problem is a multi-load truss optimisation with 68 non-symmetric topology design variables. The optimum design should be feasible considering the structural reactions subject to 8 different loading conditions. See \cite{ahrari2016improved} for details on the simulation. Table \ref{table.68bar} shows our findings in comparison with state of the art. We can see that our method obtained the best design alike to the one obtained in \cite{ahrari2016improved}, where this design incorporates 34 group bars. However, designs (b) and (c) include 37 and 39 bars, respectively.
\section{Conclusions}
In this paper we considered bilevel optimisation of topology and size of trusses subject to discrete sizes. For the lower level optimisation, we use a reliable evolutionary optimiser. For the upper level, we employ exact enumeration and novelty-driven binary particle swarm optimisation to explore the upper level. 
We also use a repair mechanism to fix the infeasible solutions in the upper level that violate truss constraints. 

In our experiments, we analysed the search space of smaller problems without randomness in the upper level. We also observed that we can find multiple distinct high-quality solutions with respect to the topology -- moreover, we have found new best solutions for 8 out of 9 test problems.  

Bilevel optimisation problems nest an optimisation problem into another where it increases the computational expense. This is the main drawback of this study. We also setup our algorithms with standard parameters. For future studies, it could be interesting (1) to investigate automated tuning of the algorithm, (2) to study this approach for large-scale trusses and (3) to improve it considering the computational expense of the problem.

\bibliographystyle{plain}
\bibliography{mybib}

\end{document}